\documentclass[final]{cvpr}

\usepackage{times}
\usepackage{epsfig}
\usepackage{graphicx}
\usepackage{amsmath}
\usepackage{amssymb}
\usepackage{tabularx}
\usepackage[pagebackref=true,breaklinks=true,colorlinks,bookmarks=false]{hyperref}

\def\cvprPaperID{****} % *** Enter the CVPR Paper ID here
\def\confYear{}

\begin{document}

%%%%%%%%% TITLE
\title{Machine Learning Approaches to Automated Flow Cytometry Diagnosis \\of Chronic Lymphocytic Leukemia.}

\author{Akum S. Kang$^{1,2}$,  Loveleen C. Kang MD$^{3, 4}$, Stephen M. Mastorides MD$^{3, 4}$, Philip R. Foulis MD$^{3, 4}$, \\Lauren A. DeLand NP$^{3}$,  Robert P. Seifert, MD$^{5}$, Andrew A. Borkowski MD$^{2,3,4}$\\
}

\maketitle

%%%%%%%%% ABSTRACT 
\begin{abstract}
Flow cytometry is a technique that measures multiple fluorescence and light scatter-associated parameters from individual cells as they flow a single file through an excitation light source. These cells are labeled with antibodies to detect various antigens and the fluorescence signals reflect antigen expression. Interpretation of the multiparameter flow cytometry data is laborious, time-consuming, and expensive. It involves manual interpretation of cell distribution and pattern recognition on two-dimensional plots by highly trained medical technologists and pathologists.  Using various machine learning algorithms, we attempted to develop an automated analysis for clinical flow cytometry cases that would automatically classify normal and chronic lymphocytic leukemia cases. We achieved the best success with the Gradient Boosting. The XGBoost classifier achieved a specificity of 1.00 and a sensitivity of 0.67, a negative predictive value of 0.75, a positive predictive value of 1.00, and an overall accuracy of 0.83 in prospectively classifying cases with malignancies.
\end{abstract}

%%%%%%%%% BODY TEXT 
\section{Introduction} 

\footnotetext[1]{College of Computing, Georgia Institute of Technology (Georgia Tech), Atlanta, GA}
\footnotetext[2]{Joint-Corresponding authors: kangakum@gatech.edu, Andrew.borkowski@va.gov}
\footnotetext[3]{James A. Haley Veterans’ Hospital, Tampa, FL.}
\footnotetext[4]{Department of Pathology and Cell Biology, University of South Florida, Tampa, FL.}
\footnotetext[5]{Department of Pathology, Immunology and Laboratory Medicine, University of Florida, Gainesville, FL.}

Flow cytometry is a technique used to detect and quantitate the physical and immunophenotypic properties of cells suspended in fluids. Flow cytometry measures multiple fluorescence and light scatter associated parameters from individual cells as they flow single file through an excitation light source$^1$. Additionally, cells are typically labeled with antibodies to detect various antigens and the fluorescence signals reflect antigen expression.  The technology converts fluorescence signals into electrical impulses, which are then digitized and processed by a computer and stored in a list mode format (Flow Cytometry Standard (FCS)). FCS files contain a matrix of expression values of all measured channels for all particles analyzed by a flow cytometer$^2$. This data generated is generally plotted in two-dimensional plots on a logarithmic scales. The regions on the plots are sequentially separated using gates, a form of subset extraction. Modern flow cytometers are able to analyze thousands of cells per second$^3$.

Clinical flow cytometry is highly suitable for diagnosis of hematopoietic disorders especially using peripheral blood, which has cells naturally suspended in a fluid base. The most common application is immunophenotyping to diagnose lymphoma and leukemias. Leukemia is a major hematological malignancy and can be acute (fast-growing) or chronic (slow-growing) and can start in myeloid cells or lymphoid cells. Chronic lymphocytic leukemia (CLL) is the most common adult leukemia in western countries. According to the National Cancer Institute’s Surveillance, Epidemiology, and End Results Program (SEER), the estimated number of new cases of CLL in the United States in 2018 was 20,940, representing about 1.2\% of new cancer diagnoses$^4$.  The disease results from the overgrowth of B lymphocytes co-expressing CD5, CD23, and low levels of surface immunoglobulin (sIg) of a single IG light (L) chain type and CD20$^4$. Peripheral blood flow cytometry is key to diagnosing CLL. However, interpretation of the multiparameter flow cytometry data generated by the flow cytometer is laborious, time-consuming and expensive as it involves manual gating, manual interpretation of cell distribution and pattern recognition on two-dimensional plots by highly trained personnel such as medical technologist and pathologist.

Augmented human intelligence (AHI) and artificial intelligence (AI) tools are being increasingly incorporated in various fields of medical practice for both diagnosis and patient management. With increasing understanding of the disease processes and increased testing platforms availability, the data generated has also significantly increased. Usage of  AHI tools like machine learning (ML) and deep learning algorithms in medicine is being increasingly evaluated to analyze the vast amount of data available and adequately incorporate in medical practice$^{6,7}$.  Machine learning has been applied in flow cytometry including Critical Assessment of Population Identification Methods(FlowCAP)$^8$, in which automated  algorithms are generated to reproduce manual gating. However, these methodologies also are dependent on extrapolating gates and user bias. Given the limitation of some of the tried methods, we attempted to develop and automated analysis for clinical flow cytometry cases that would allow for automatic classification of normal and chronic lymphocytic leukemia cases .

\section{Methods and Materials} 
Institutional review board approval was granted for this study.  Deidentified flow cytometry data from the lymphoma panel and the associated diagnostic flow cytometry reports from peripheral blood specimens from March 2017 to November 2019 were obtained. Blood samples were collected in an EDTA preservative. The panel is a 4 tube 10-color laboratory-developed test that contains antibodies to kappa light chain (polyclonal and monoclonal), lambda light chain (polyclonal and monoclonal), CD45, CD2, CD3, CD4, CD5, CD7, CD8, CD10, CD11c, CD16, CD19, CD20, CD22, CD23, CD25, CD30, CD38, CD52, CD56, CD57, CD103, CD200, FMC-7, HLA-DR, T-cell receptor a/b, and T-cell receptor g/d.  Specimens were run on BD FACSCanto Clinical Flow Cytometery System.
10 fluorochromes per tube are used to measure specific antibodies. From each tube, 13 channels of data are collected (forward scatter-area (FSC-A), forward scatter height (FSC-H), side scatter area (SSC-A), and 10 fluorescence signals). Together, fluorescence signals from fluorochrome-labeled antibodies, forward and side-scatter measurements collected from the 4 tubes as FCS files are analyzed. The median number of events collected from this panel per run was 77,416, with a range of 15,574 to 904,338. CLL cases include normal CLL and monoclonal B cell lymphocytosis of CLL immunophenotype, which is a distinction made on a number of circulating monoclonal cells in the peripheral blood. The decision was made to have one class of “positive” cases, grouping CLL and MBCLL together.

Table 1 includes a breakdown of the types of cases in the study.
\begin {table}
\caption {Characteristics of Patient Cases} \label{tab:title} 
\begin{center}
\begin{tabular}{||c c ||} 
 \hline
 \textbf{Case Type} & \textbf{No.(\%) of cases} \\ [0.5ex] 
 \hline\hline
 Normal & 53 (45.69)  \\ 
 \hline
 CLL & 44 (37.93) \\
 \hline
 MBCLL & 19 (16.38) \\ [1ex] 
 \hline
\end{tabular}
\end{center}
\end {table}

Analysis was performed in Python 3.7 using a Macbook Pro with 8GB of RAM and a dual-core Intel Core i5 processor, and also tested on an Ubuntu 18.04 LTS machine with 32GB RAM, an Intel Core i7-9700K processor, and an Nvidia RTX 2080 graphics card.  The FCS files containing events from the 116 cases (464 tubes) were compressed into a 2-dimensional data matrix.  The first 384 events (5000 array elements) from each channel were excluded, and the following 769 events (10,000 array elements) from each of the 13 channels were concatenated together to obtain an array of size 130,000 (representing 10,000 events) for each sample.  Next, the samples corresponding to a specific patient were concatenated together to obtain a 1-dimensional array of size 520,000 (representing 40,000 events) per case.  After the data cleaning, a total of 4,640,000 events were collected from 116 peripheral blood cases.  The cases underwent a standard 80/20 split where a random selection of 92 cases was used for training, and 24 were used for testing.

The rows of the data matrix were used to train various machine learning models.  First, data was fed into a random forest classifier with a Gini Impurity criterion, a maximum depth of 2, and 100 estimators from sklearn 0.22$^9$.  A neural network with 3 dense layers and a dropout layer with a rate 0.5 from TensorFlow 2.4$^{10}$ was also trained with the data. Additionally, a 1-dimensional convolutional neural network with 2 convolutional layers and a pooling layer added onto the vanilla neural network was trained. Finally, a gradient boosting classifier with 100 trees and a maximum depth of 3 from XGBoost 1.4.2$^{11}$ was trained on the patient data.  The random forest and gradient boosting classifiers were chosen because of the historically good performance of decision trees on flow cytometry data in relation to hematologic malignancies$^{12}$.  The Neural Network and CNN were chosen because of their wide range of applications in medical AI and other fields, in addition to their specific performance in classifying flow malignancies$^{13}$.

Evaluation of model performance stability for the XGBoost model was done using a 10-fold cross-validation with 80/20 splits on the entire dataset, again 92 cases for training and 24 for testing each time.  Results for all cross-validation runs were averaged together, and the standard deviation was computed.  The cross-validation accuracy was used to ensure the stability of the models across different train/test splits.

\section{Results}

\subsection{Time}
Data transformation and processing on 4 million events took 3:20 minutes.  This included removal of the first 384 events and concatenation of the next 769 events across the 13 channels and 4 tubes per patient.  Training the RFC took 32 seconds, training the vanilla neural network took 2:50 minutes, training the CNN took 3:19 minutes, and training the XGBoost classifier took 3:07 minutes.  Subsequent predictions for each model took less than 1 second each.

\subsection{Accuracy}
Using just the B-cell panel, the XGBoost classifier achieved a specificity of 100.0\% and a sensitivity of 67.67\%, a negative predictive value of 75.00\%, a positive predictive value of 100.00\%, and an overall accuracy of 83.33\% in prospectively classifying cases with malignancies.  These malignancies included CLL and MBCLL.  The RFC achieved an overall accuracy of 79.16\%.  The neural network achieved an overall accuracy of 51.09\%, and the CNN achieved an accuracy of 45.65\%.  The bulk of the subsequent analysis will be focused on the XGBoost classifier because it achieved the best results. Figure 1 shows a ROC (receiver operating characteristic) curve for the XGBoost classifier's predictions on the test cases.

\begin{figure}[htp]
    \centering
    \includegraphics[width=8cm]{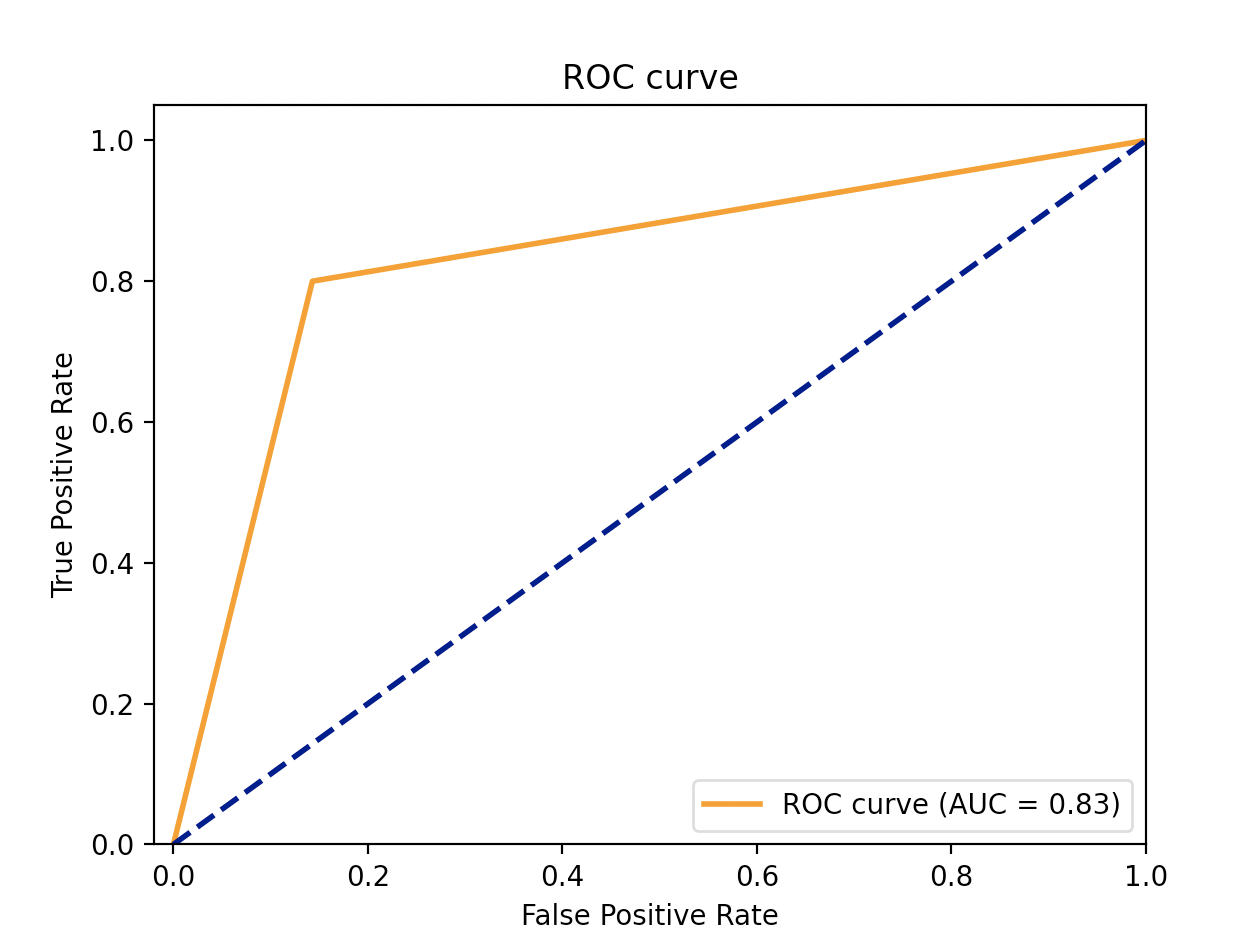}
    \caption{ROC curve for classification of all cases (CLL, MBCLL, and cases without diagnosed abnormality).}
    \label{fig:ROC}
\end{figure}

\subsection{Analysis of cases misclassified by XGBoost}
Out of the 24 testing cases, 20 were correctly classified.  There were 4 false-negative cases, which were misclassified as negative. All four cases had the lymphoma cells in the range of what is classified as monoclonal B cell lymphocytosis of CLL immunophenotype (MBCLL) (less than 5 X 10 E9/L cells). These four cases had an absolute count of lymphocytes with CLL immunophenotype ranging from 1.47 to 3.49 x 10E9/L.

\subsection{XGBoost Classifier Stability}
10-fold cross-validation of the Gradient Boosted classifier showed similar results with less than 10\% standard deviation in the accuracy.  Additionally, it showed an F1 score of 79.6\% with less than 11\% standard deviation.  As expected, repeated runs of the XGBoost model showed less than 5\% difference in accuracy across all classes.

\section{Discussion}

\subsection{Analysis of Model Performance}
The XGBoost model was able to classify cases with the highest accuracy out of all classifiers.  Recent studies comparing Gradient Boosting to Deep Learning models suggest that when the data is heterogeneous, Gradient Boosting outperforms simple Neural Nets, or even more advanced architectures like ResNets$^{14}$.  For our study, all 13 channels from 4 patient were combined together to create one row of data points per patient.  This heterogeneous combination of a variety of channels, including fluorescence signals, forward scatter, and side scatter parameters, could potentially explain the superiority of Gradient Boosting when compared with Neural Network architectures on our dataset.  Additionally, Gradient Boosting outperformed a Random Forest model.  Even though both architectures are ensemble models, a Gradient Boosting architecture fits each weak classifier to the residual of the previous classifier's performance, leading to a model with higher accuracy and greater stability$^{11}$.  This could also explain why the cross-validated stability of the model was very close to the reported accuracy from just one testing run.

Additionally, all our models were limited in the small number of cases available for training and testing.  Various statistical studies show that ideally, at least 75-100 testing data points are needed for tabular data modeling$^{15}$, which is much greater than the 24 samples used in our study.  Additionally, increasing the amount of available training data may help the performance of neural network architectures trained on TensorFlow $^{10}$.

\subsection{Pathological Analysis}

The Gradient Boosting model was able to identify all the cases of chronic lymphocytic leukemia with high accuracy, and the analysis was made in less than a minute after the training was done. This lends the possibility of an expedited review by the pathologist and thus an alert to the treating physician. Going further, this model has the potential to be applied to other disease panels (acute leukemia, plasma cell, myeloid panel etc.) for expedited initial classification, expedited review and thus leading to prompt initiation of treatment, which is critical in some acute processes.
 
There were a few cases of monoclonal B-cell lymphocytosis of CLL, which were misclassified by our model as normal. However, our model was limited by a small number of cases in the training set. This could probably be improved by adding more cases to the training set; however, smaller laboratories are limited by an indigenous number of cases, which may be overcome by collaborating with additional laboratories. Collaboration with other laboratories could introduce inter-instrument variation, which may require streamlining the current data processing pipeline to transform all inputs into a consistent format. Also, increasing the training set may improve performance for classifying normal cases.
 
The results from this study are encouraging for possible future use of machine learning in clinical and diagnostic flow cytometry. Initially, this technique may be applied within clinical laboratories for retrospective review of all cases for quality control followed by peer review of false positive and false negative cases, instead of the currently used manual review of only randomly selected cases. Prospectively the methodology may generate a preliminary interpretation of the data into normal versus abnormal and subsequently into diseases for the pathologist to review. As we develop more experience with this methodology, there is a potential of auto verification of the normal cases as we do in other aspects of clinical testing. Machine learning in flow cytometry has potential for research implications as well. Computationally analyzing our expanding repertoire of flow cytometry data by machine learning opens new avenues for identifying differences between normal and abnormal cell populations and may lead to new discoveries.

\section{Conclusion}
Given the data constraints, the Gradient Boosting classifier performed well on the testing dataset.  To further expand this study, we recommend collaboration with other laboratories to increase the dataset size and streamline the data processing pipeline.  Acquiring more data will open up the possibility of training more accurate Deep Learning models to classify CLL.  Additionally, we recommend this study be expanded to immunophenotyping of other hematological malignancies.

\section{Funding}
This material is the result of work supported with resources and the use of facilities at the James A. Haley Veterans’ Hospital

\section{References}
1.	Shapiro HM (2003). Practical flow cytometry (4th ed.). New York: Wiley-Liss. ISBN 978-0-471-41125-3.

2.	Spidlen J, Moore W, Parks D, Goldberg M, Bray C, Bierre P, Gorombey P, Hyun B, Hubbard M, Lange S, Lefebvre R, Leif R, Novo D, Ostruszka L, Treister A, Wood J, Murphy RF, Roederer M, Sudar D, Zigon R, Brinkman RR. Data File Standard for Flow Cytometry,Version FCS 3.1. Cytometry Part A 2010;77A:97–100.

3.	Michael Brown, Carl Wittwer, Flow Cytometry: Principles and Clinical Applications in Hematology, Clinical Chemistry, Volume 46, Issue 8, 1 August 2000, Pages 1221–1229, https://doi.org/10.1093/clinchem/46.8.1221

4.	SEER Cancer Stat Facts: Chronic Lymphocytic Leukemia, National Cancer Institute, Bethesda, MD; https://seer.cancer .gov/statfacts/html/clyl.htm

5.	Nicholas Chiorazzi, Shih-Shih Chen, and Kanti R. Rai. Chronic Lymphocytic Leukemia. Cold Spring Harb Perspect Med 2021;11:a035220

6.	Beam A, Kohane I. Big data and machine learning in healthcare. JAMA. 2018; 319: 1317‐ 1318

7.	Topol E. High‐performance medicine: the convergence of human and artificial intelligence. Nat Med. 2019; 25(1): 44‐ 56

8.	Aghaeepour N, Finak G, Hoos H, et al; FlowCAP Consortium; DREAM Consortium. Critical assessment of automated flow cytometry data analysis techniques. Nat Methods. 2013;10:228-238.

9. Pedregosa F, Varoquaux G, Gramfort A, et al. Scikit- learn: machine learning in Python. arXiv e-prints. 2012. arXiv:1201.0490.

10. Abadi, Ashish Agarwal, Paul Barham, et al. TensorFlow: Large-scale machine learning on heterogeneous systems,
2015. Software available from tensorflow.org.

11. Chen, T, Guestrin, C. (2016). XGBoost: A Scalable Tree Boosting System.. In B. Krishnapuram, M. Shah, A. J. Smola, C. Aggarwal, D. Shen,  R. Rastogi (eds.), KDD (p./pp. 785-794), : ACM. ISBN: 978-1-4503-4232-2

12. David P Ng, MD, Lauren M Zuromski, MS, Augmented Human Intelligence and Automated Diagnosis in Flow Cytometry for Hematologic Malignancies, American Journal of Clinical Pathology, Volume 155, Issue 4, April 2021, Pages 597–605, https://doi.org/10.1093/ajcp/aqaa166

13. Li, Y., Mahjoubfar, A., Chen, C.L. et al. Deep Cytometry: Deep learning with Real-time Inference in Cell Sorting and Flow Cytometry. Sci Rep 9, 11088 (2019). https://doi.org/10.1038/s41598-019-47193-6

14. Yury Gorishniy, Ivan Rubachev, Valentin Khrulkov, Artem Babenko. Revisiting Deep Learning Models for Tabular Data. https://arxiv.org/pdf/2106.11959v1.pdf

15. Claudia Beleitesa, Ute Neugebauer Thomas Bocklitzc, Christoph Krafft, Jürgen Poppa,b,c. Sample Size Planning for Classification Models. https://arxiv.org/pdf/1211.1323.pdf.

\section{Appendices}
\subsection{Experimental Details}
\subsubsection{Source code}
Publicly available: \href{FlowAI}{https://github.com/kangakum36/FlowAI}

\end{document}